\documentclass[runningheads]{llncs}
\usepackage[T1]{fontenc}
\usepackage{graphicx}
\usepackage{hyperref}
\renewcommand\and{\\[\baselineskip]}

\begin{document}
\title{Multi-Module System for Open Domain Chinese Question Answering over Knowledge Base}
\titlerunning{Multi-Module System for Open Domain KBQA}
%
\author{Yiying YANG\inst{1} \and Xiahui HE\inst{2} \and Kaijie ZHOU\inst{1} \and Zhongyu WEI\inst{2}}
\institute{Ping An Life Insurance of China, Shanghai 200120, China \\
\email{\{yangyiying283,zhoukaijie002\} @pingan.com.cn}\\ \and
 Fudan University, Shanghai 200433, China \\
\email{\{18210980050,zywei\} @fudan.edu.cn}\\}
\maketitle              
\begin{abstract}
For the task of open domain Knowledge Based Q\&A in CCKS2019, we propose a method of combining information retrieval and semantic parsing, which extracts the topic entity and the most related relation predicate from a question and transforms it into a Sparql query statement. Our method obtained an F1 score of 70.45\% on the test data.
\end{abstract}

\section{Introduction}
We introduce an open domain question answering system based on Chinese knowledge graph in this paper. We analyze the questions and find that most of the answers to the questions are within two hops. Therefore, we only solve the problem within two hops to reduce the complexity of the system. The system consists of a topic entity selection module, a relationship recognition module and an answer selection module. Firstly, we construct a scoring mechanism to select the core entity of the question; Then we score the relationship in the two-hop subgraph of the topic entity; Finally, we build a classier to judge whether a question is simple or complicated, so that we can choose the final relationship and generate sparql query. 

\section{Related Work}
There are two main approaches in Knowledge Graph based Question Answering(KBQA) : semantic parsing based and retrieval based. 

Semantic Parsing based approach is a linguistic method that transforms natural language into logic forms and queries them in the knowledge graph through corresponding semantic representations, such as lambda-Caculus, to arrive at an answer. Semantic parsing-based methods, including semantic parsing based on Lambda Dependency-Based Compositional Semantics (Lambda-DCS)~\cite{liang2013lambda,berant2013semantic}, semantic parsing based on Combinatory Categorical Grammars (CCG), semantic parsing based on Neural Machine Translation(NMT), and semantic parsing based on deep learning~\cite{yih2015semantic} released by Microsoft in 2015.

Retrieval based approach could be regarded as a sorting algorithm for the answer: given the input question Q and the knowledge graph KB, by scoring and sorting the entities in the KB, the entity or entity set with the highest score is selected as the answer.
It mainly includes feature-based method \cite{yao2014information}, extracting features from the input question Q and the answer candidate A, generating feature vectors, and training the classifier; vector-representation based method \cite{bordes2014question}, the input question Q and the answer candidate A are represented as two vectors (distributed embedding) respectively, and vector distance is calculated for scoring; CNN network based method \cite{dong2015question}, the feature extraction is performed by a convolutional neural network; Gated-GNN based method\cite{sorokin2018modeling}, etc.

We combine the above two methods. On the one hand, we use the retrieve based method to sort KB relationships and entities, and on the other hand, we use the most related relationship and entity to generate the sparql statement to query the final answer.

\section{The Proposed Model}

\begin{figure} 
\centering
\includegraphics[width=4.5in]{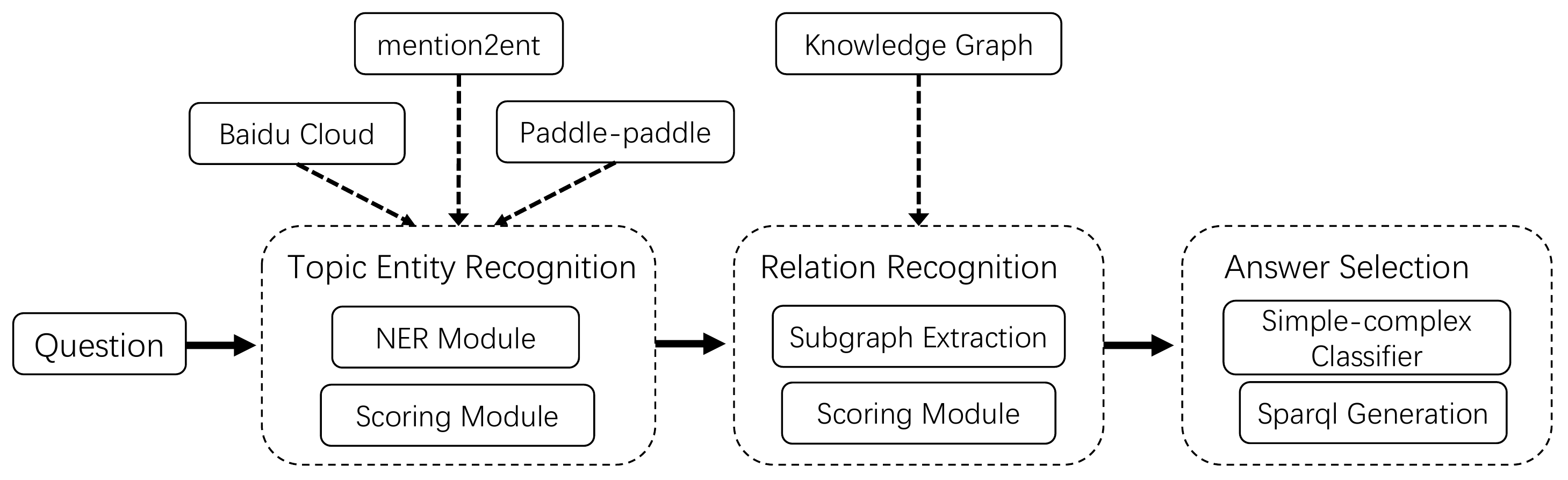}
\caption{The overall structure of our model.}
\label{fig.model}
\end{figure}

Our model is mainly divided into three parts, namely \textbf{Topic Entity Recognition}, \textbf{Relation Recognition} and \textbf{Answer Selection}. The overall model is shown in Figure \ref{fig.model}. 

After entering a question, the model first finds the topic entity in the sentence. Here we used the Entity-mention file provided by the contest organizer and some external tools such as paddle-paddle. Then in the relationship recognition module, the relationship of the question (also called the predicate) is found by extracting the subgraph of the topic entity in the knowledge graph. The ranking of all relationships is obtained by a similarity scoring model. Finally, in the answer selection module, according to the simple-complex problem classifier and some rules, the final answer is obtained.

\subsection{Topic Entity Recognition}
Topic entity is the core entity of the corresponding KB query. Since most of the problems are within two hops, so we only need to find a core entity and then search for its subgraph to find the answer of the question. The extraction of the topic entity is divided into two parts. Firstly, all the entities are extracted through the NER module. Then, the entities are scored by some rules and the entity with the highest score is selected as the topic entity.

\subsubsection{NER Module}
We chose the pre-trained NER models released by Baidu Cloud and Paddle-paddle. These two models have their own preponderance in entity recognition, so we decide to combine the results of both. Unlike the common NER, we don't just extract the entities labeled 'LOC', 'ORG', 'PER' and 'TIME'. Since the entity in question is not always one of the four, we also extract  'n' (noun), 'nr' (person name), 'ns' (place name), 'nt' (institution name), 'nw' (work name). Besides, considering that there are many financial problems in this data set,  we have compiled a small dictionary of vocabulary in the financial field on the public website in order to ensure that the model has strong adaptability.

\subsubsection{Stop-words}
In Chinese, many relationships exist in the form of nouns, so they are extracted in the NER module, such as `creator' and `inventor'. However, our topic entity should not use relationship as the core node, so we maintain a stop-words dictionary, which records the relationship in the form of nouns. Considering that some of the words also have the possibility of being a real entity,  we set a rule that when a sentence has not been checked by any entity after passing the previous NER module , we reconsider detecting the words in the dictionary. 

\subsubsection{Scoring Module}
After the NER module we get a series of entities, we select all possible entities according to the correspondence in the entity-mention dictionary. If an entity does not appear in the dictionary, we further query it in the knowledge base, and if it still does not, discard it. Then we constructed a scoring strategy for all the candidate entities generated above. The specific score rules are as follows:

\textbf{Score1} The Length of Entity. An entity with a longer length is usually given a higher score.

\textbf{Score2} The Out-degree of Entity in the KB. An entity with higher out-degree is more likely to be the topic entity. Meanwhile, it is more efficient to query the nodes pointed to other entity in the knowledge base than to be pointed. 

\textbf{Score3} The Distance Between the Entity and Interrogative Word. We define ['who','what','where','how','how much','how many'] as interrogative words. An entity gets higher score if more close to the above words.

\textbf{Score4} Char Overlap between Entity and Question. The more overlap chars shared between entity and question, the bigger probability that the entity be a topic entity.

\textbf{Score5} Word Overlap between Entity-mention and Question. If an entity and its corresponding mention have more overlap with the problem, it is more likely to be the topic entity.

\textbf{Score6} NER Label of Entity. If an entity is recognized as special noun like 'ORG', it has more probability to be a topic entity than simple noun. Additionally, if an entity is a person name or financial related, we assign it higher score.

\textbf{Score7} Similarity between Entity-mention and Question. We simply fine-tune bert model to measure how well an entity matches a question. Matching rate is the corresponding score.

\begin{figure} 
\centering
\includegraphics[width=4in]{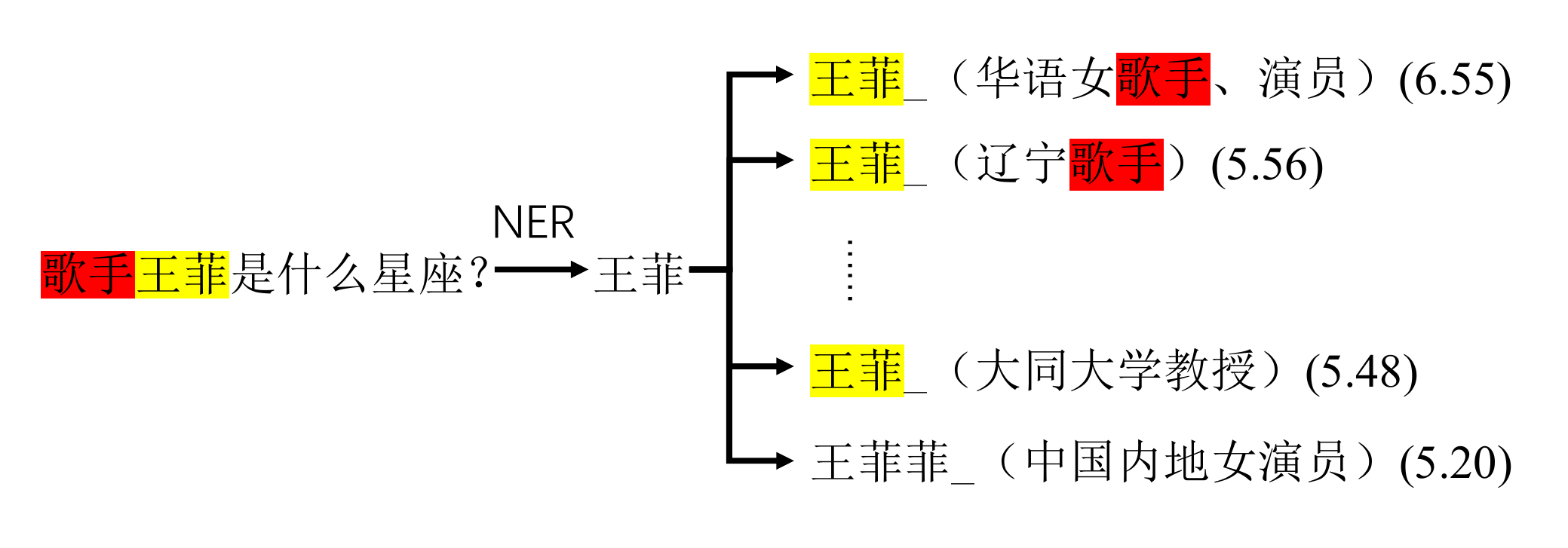}
\caption{An example of topic entity recognition}
\label{fig.ner}
\end{figure}

After calculating the above scores, we normalize and add them, the entity with the highest score is selected as the topic entity. Figure \ref{fig.ner} shows the flow of extraction. The scores of the first two entities are higher because they capture more keywords in the question and the reason why the first entity is higher than the second is that its out-degree is much higher, which means that it is more well-known. 

\subsection{Relation Recognition}

The relationship in the question generally refers to the predicate in a sentence, and also the relationship between two entities in the knowledge base. The correct relationship should be the relationship between the topic entity and the answer entity.

\subsubsection{Subgraph Extraction}

We mainly use retrieval based method in recognizing relationships. By observing the dataset, we found that among the 2,298 questions in the training set, there were 1,160 questions that could be solved through one-hop in the knowledge graph, and 912 questions through two-hop, accounting for 90.17\% of all data. Thus, We only consider the problem within two-hop in later experiments.

For each topic entity, we extract the subgraphs within its two-hop in the knowledge base, and each relationships in the subgraph could be the target relationship.

\subsubsection{Scoring Module}

After getting all the candidate relationships, we constructed a scoring strategy. The specific score rules are as follows:

\textbf{Score\_relation\_similarity} Similarity between Relation and Question. We use the \textbf{BERT-Base, Chinese}~\cite{devlin2018bert} pre-trained model to initialize word vectors. Then fine-tune it through a similarity model~\cite{zhou2017InsunKBQA}. Question and relationship are first fed into the word embedding layer, mapping words to a fixed-dimensional word vector. It is then sent to the biLSTM layer, whose output is averaged at each step. The obtained result is sent to a fully connected layer, and finally the semantic embedding is obtained. The cosine similarity of the two semantic embeddings is calculated to represent the similarity between the question and relation. 

\begin{figure} 
\centering
\includegraphics[width=4.5in]{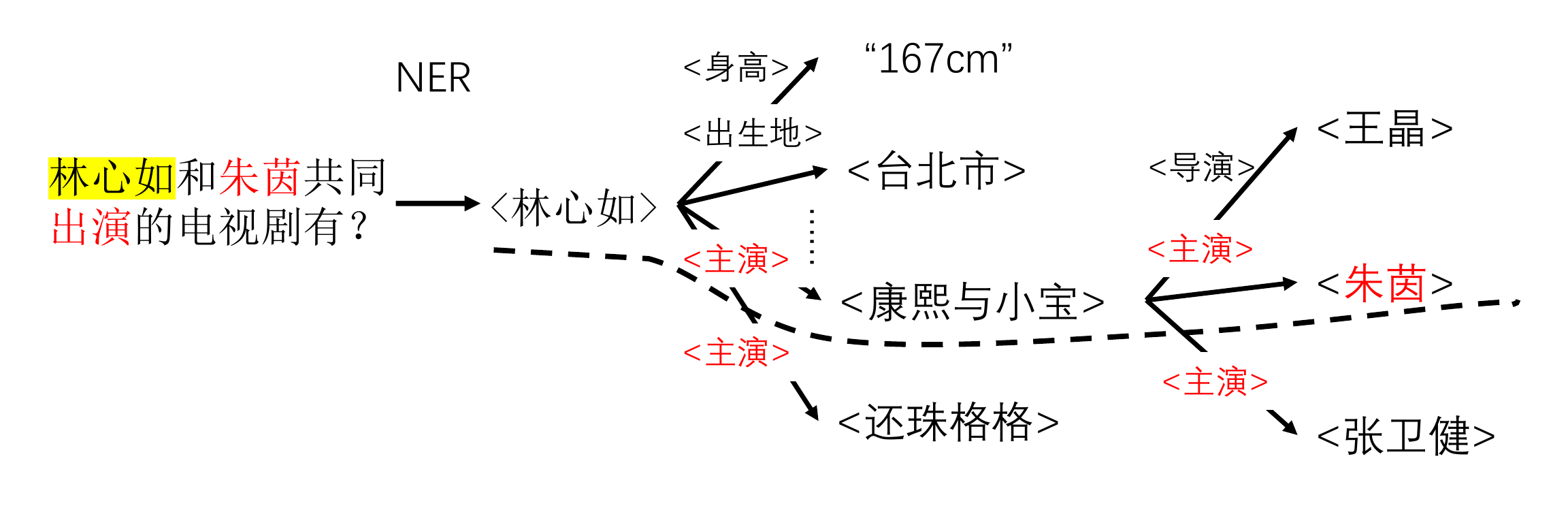}
\caption{An example of relation recognition}
\label{fig.rel}
\end{figure}

\textbf{Score\_object\_similarity} Similarity between Object and Question. This model is similar to the above relation similarity model. Although the similarity between relation and question is the main indicator of evaluation, the similarity between object and question also plays a supporting role. For example, the question "What TV series did actor A and actor B play together?", if actor A is recognized as the topic entity, and actor B appears as an object of a two-hop relation(actor A ==act in==> TV serie X <==act in== actor B), the score of this relation chain should be higher than other relations. An example is shown in figure \ref{fig.rel}.

\textbf{Score\_char\_overlap} Char Overlap between Relation and Question. The more overlap chars shared between entity and question, the bigger probability that the relation be a correct relation. This score is added primarily to prevent the model from relying too heavily on the bert similarity model.

The final score of the relationship is the weighted addition of the above scores. 

\subsection{Answer Selection}

After getting the topic entity and relationship scoring of the question, we need to generate the final sparql query and find the answer from the knowledge graph.

\subsubsection{simple-complex Question Classifier}

As mentioned before, we learned a classifier since we only consider one-hop (simple) and two-hop (complex) problems. Before performing the final sparql generation, we use the classifier to determine whether it is a simple or complex problem. If it is a simple problem, select the highest one-hop relationship as the answer. Conversely, if it is a complex problem, select the highest two-hop relationship.

We use a bert classifier to implement this classification model.

\subsubsection{Sparql Generation}

We need to generate the corresponding sparql statement based on the topic entity and the most relevant relation, and find the result from the knowledge graph.

\begin{figure} 
\centering
\includegraphics[width=5in]{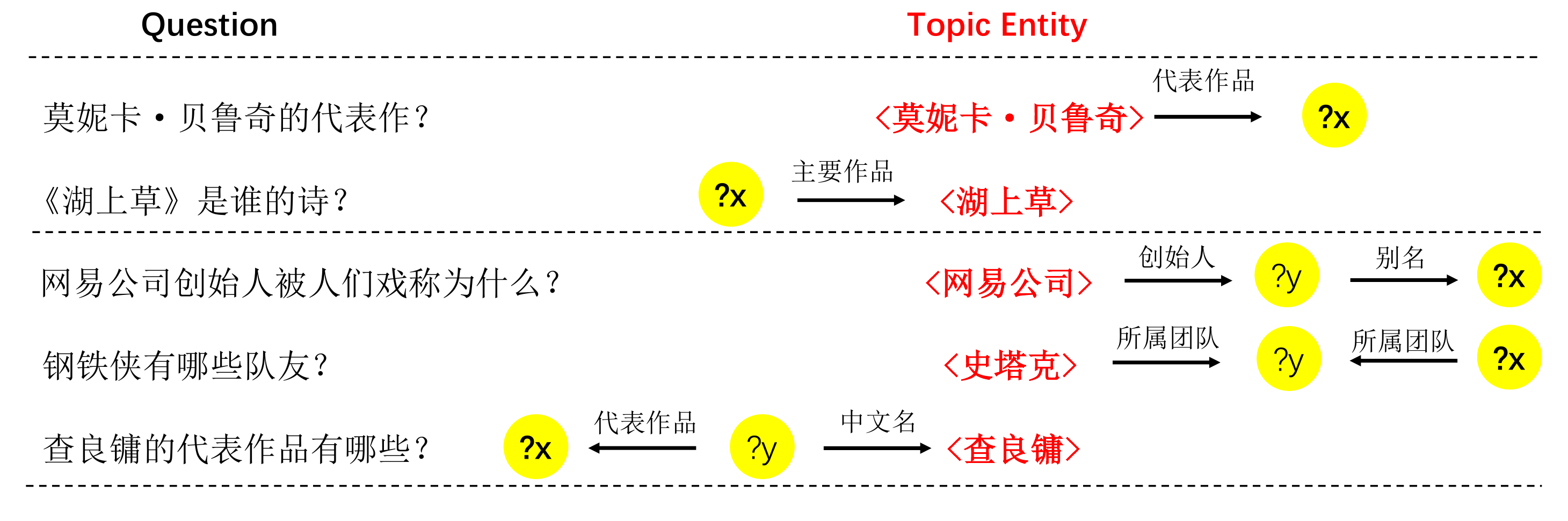}
\caption{Relation Structure}
\label{fig.structure}
\end{figure}

We consider the relationship of the five structures shown in Figure \ref{fig.structure}. For the question "What is the representative work of Monica Bellucci?", we found the topic entity <Monica Bellucci>, it is classified as a simple one-hop question, the highest score relation is <representative work>, and the sparql statement is "select ?x where {<Monica Bellucci> <representative work> ?x}".
 
In addition, we have added some other rules, such as:

(1) When there is a gender-related vocabulary in the question, add “{?z <gender> <male/female>}” to sparql as restriction.

(2) When time appears in the question, it is unified into the format of “YYYY-MM-DD” (consistent with the knowledge graph).

(3) When the two-hop object appears in the sentence, select the intermediate result ?y is the final result. Still consider question “What TV series did actor A and actor B play together?”, the highest score is “actor A ==act in==> TV serie X <==act in== actor B”, but ?y(TV serie X ) rather than ?x(actor B) is the correct result. Thus, the sparql should be “select ?y where {<actor A> <act in> ?y. <actor B> <act in> ?y.}”

\section{Experiments and Results}

\subsection{Dataset}

We use CCKS 2019 dataset to evaluate our approach. The dataset is published by the CCKS 2019 task 6, which includes a knowledge base, an entity-mention file, and Q\&A pairs for training, validation, and testing. The knowledge base has more than 30 million triples (We use gstore\footnote{https://github.com/pkumod/gStore} to manage the knowledge base), the training set has 2298 question and answer pairs, the dev set has 766 questions, and the test set has 766 questions. Since we don't have the correct answer to the dev set, in order to evaluate the model performance during the experiments, we randomly selected \textbf{100} Q\&A pairs from the training set as the real development set.

\subsection{Topic Entity Recognition}

We test the effects of different scores in the scoring mechanism. Table~\ref{tab1} shows some results. Baseline refers to the sum of the scores excluded similarity score and out-degree score. We find that the impact of out-degree score is mainly reflected in confused entities. As shown in Figure \ref{fig.ner}, there are actually two singers named 'Wangfei'. At this time, we need to combine common sense to select the more likely one. Generally speaking, the more well-known entities correspond to more information (that is, the introduction more detailed). The score of similarity gives a higher weight to the entity more relevant to the sentence.

\begin{table}
\caption{The effect of different scores on extracting topic entity}\label{tab1}
\centering
\begin{tabular}{|l|c|}
\hline  
Score&accuracy\\
\hline  
baseline&90.3\%\\
baseline+out-degree&92.3\%\\
baseline+similarity&93.8\%\\
baseline+out-degree+similarity& \bf 95.6\%\\
\hline 
\end{tabular}

\end{table}

\subsection{Relation Recognition}

When constructing the training set for the similarity scoring model, since the positive samples are significantly smaller than the negative samples, we performed 5 oversamplings on the positive samples and 5 randomly selected training sets from all negative samples of each data. The positive and negative relationship of the one-hop relationship is easier to understand. The positive sample of the two-hop relationship is the concatenation of the correct one-hop and two-hop relationship, and the wrong relationship is not put into the negative sample because it may interfere with the one-hop relationship score. In addition, entities in all questions are replaced with <e> in order to reduce entity interference. We tried several common models, the results of relation scoring model is shown in table \ref{tab.rel}. Bert model has the highest accuracy of 95.7\%.

\begin{table}
\caption{Results of Relation Recognition. Note that the accuracy here is for relational samples rather than Q\&A samples.}\label{tab.rel}
\centering
\begin{tabular}{|l|c|}
\hline
Model & Accuracy \\
\hline
Similarity-textRNN & 85.4 \%   \\
Similarity-textCNN & 87.3 \%  \\
Similarity-Bert & \bf 95.7 \%  \\
\hline
\end{tabular}
\end{table}

\subsection{Answer Selection}

The simple-complex model is a simple binary classifier, it has an accuracy rate of 91\%. Final Answer Selection results are shown in table \ref{tab.ans}. 
We evaluated the model using accuracy indicator. The baseline model, which is the bert relation similarity model mentioned above, has an accuracy of 68\% over 100 dev data. After adding the object similarity score and sparql rules, the accuracy is increased to 75\%.

Since the correct answer to test set has not yet been released, we are unable to verify the accuracy of each model. According to the final version submitted on the website, our model has a F1-score of 70.45\% in test set.

\begin{table}
\caption{Results of Final Answer Selection. The star(*) number waits for the correct answer to be published.}\label{tab.ans}
\centering
\begin{tabular}{|l|c|c|}
\hline
Model & Dev Set (accuracy) & Test Set (F1-score) \\
\hline
baseline & 68\%  & *\\
baseline+object & 72\%  & * \\
baseline+object+rules & \bf 75\% & 70.45\% \\
\hline
\end{tabular}
\end{table}

\section{Conclusion}

We introduce an open domain question answering system based on Chinese knowledge graph in this paper. The system consists of a topic entity selection module, a relationship recognition module and an answer selection module. Our method obtained an F1 score of 70.45\% on the test data.

%
%
%
\bibliographystyle{splncs04}
\bibliography{mybibliography}

\end{document}